\documentclass{ar-1col}
\usepackage[numbers]{natbib}
\usepackage{url}
\usepackage[caption=false]{subfig}

\usepackage{graphicx}
\usepackage{wrapfig}

\usepackage[draft]{fixme}
\fxsetup{inline,nomargin,theme=color}

\usepackage{amsmath}
\usepackage{amssymb}
\newcommand{\mathbold}[1]{{\ensuremath{\boldsymbol{\mathbf{#1}}}}}

\usepackage[linktoc=all]{hyperref}
\usepackage[all]{hypcap}

\usepackage[nameinlink]{cleveref}

\newcommand{\g}{\,|\,}

\newcommand{\E}{\mathbb{E}}

\newcommand{\mbc}{\mathbold{c}}

\newcommand{\mbx}{\mathbold{x}}

\newcommand{\mbz}{\mathbold{z}}

\newcommand{\mbtheta}{{\mathbold{\theta}}}

\jname{Xxxx. Xxx. Xxx. Xxx.}
\jvol{AA}
\jyear{2021}
\doi{10.1146/((please add article doi))}

\begin{document}

\markboth{Chen et al.}{Probabilistic machine learning for Healthcare}
    
\title{Probabilistic Machine Learning for Healthcare}

\author{Irene Y. Chen,$^{1,*}$ Shalmali Joshi,$^{2,*}$ Marzyeh Ghassemi,$^{2,3}$ and Rajesh Ranganath$^{4,5,6}$
\affil{$^1$Electrical Engineering and Computer Science, Massachusetts Institute of Technology, Cambridge, MA 02139, USA; email: iychen@mit.edu}
\affil{$^2$Vector Institute, Toronto, Ontario, Canada; email: shalmali@vectorinstitute.ai}
\affil{$^3$Department of Computer Science, University of Toronto, Toronto, Ontario, Canada}
\affil{$^4$Department of Computer Science, Courant Institute, New York University, New York, New York 10012, USA}
\affil{$^5$Center for Data Science, New York University, New York, New York 10012, USA}
\affil{$^6$Department of Population Health, New York University Grossman School of Medicine, New York, New York 10012, USA}
}

\begin{abstract}
Machine learning can be used to make sense of healthcare data.
Probabilistic machine learning models help provide a complete picture of 
observed data in healthcare. 
In this review, we examine how probabilistic machine learning can advance healthcare. We consider challenges in the predictive model building pipeline where probabilistic models can be beneficial including calibration and missing data. Beyond predictive models, we also investigate the utility of probabilistic machine learning models in phenotyping, in generative models for clinical use cases, and in reinforcement learning. 
\end{abstract}

\maketitle

\tableofcontents

\section{INTRODUCTION}

Recent advances in healthcare---from tracking patient health with mobile phones \cite{steinhubl2013can}, to predicting peripheral vascular disease from retinal images \cite{poplin2018prediction}, to detecting sepsis early in intensive care units \cite{nemati2018interpretable}, to summarizing medical records to reduce the digital burden on clinicians \cite{pivovarov2015automated}---have made use of artificial intelligence and machine learning. The power of machine learning is rooted in both the assumptions encoded in a model
and the troves of healthcare data used for training. 

Abstractly, healthcare data can be thought of as coming from a \textit{probability distribution} called the \emph{data generating distribution}. It is through this data generating distribution that learning and evaluating a machine learning model can be formalized. For example, the error of a machine learning model on a test set, i.e., data that is not used during training, measures how well this model performs on new samples from the data generating distribution.
In many uses of machine learning models in healthcare, the data generating distribution does not need to be made explicit. For instance, making predictions or finding regression coefficients may not require consideration of the data generating distribution.

We illustrate why a probabilistic machine learning model is helpful in healthcare. Imagine modeling the survival time for melanoma patients based on stage and demographic information. If two observed stage, demographic pairs have the same average survival time, but the variance in survival differs,
the average survival time can mislead planning.
A probabilistic model of survival would instead provide a more holistic view by returning a distribution over survival times given the stage and demographic information. Such a probabilistic model would enable both the patient and the provider to better plan for the future by incorporating uncertainty into the decision making process. 
\begin{marginnote}[]
    \entry{Probability distribution}{Function describing the random process of possible outcomes, denoted $p(\cdot)$}
\entry{Data generating distribution}{The probability distribution from which observations are sampled}
\entry{Machine learning model}{An algorithm that has been trained on data for a task}
\entry{Algorithm}{A finite sequence of well-defined instructions used to solve a class of problems}
\end{marginnote} 
For example, a melanoma patient with high certainty about a small survival time might choose to make specific life adjustments. 

The probabilistic perspective can aid the entire machine learning model development and maintenance pipeline. Missing values can be mollified by probabilistic models. Censoring of labels, such as survival time, can be addressed through probabilistic models.
The probabilistic view can also aid in maintaining deployed models by detecting shifts in a data generating distribution over time. 

Probabilistic machine learning models have a myriad of practical uses in healthcare. A type of probabilistic model called a latent variable models can be used for phenotpying.
Probabilistic models can also be used for simulation, which has seen success
in scientific discovery like in drug development~\citep{segler2018generating}. The probabilistic approach
has shown promise in learning policies for diabetes and sepsis management, among other applications that use reinforcement learning~\cite{foxdeep,komorowski2016markov}.
However, estimating probabilistic models comes with a cost. Probabilistic models can require more mental and computational labor than non-probabilistic approaches, though this labor is being reduced through the introduction of probabilistic building blocks in common machine learning toolkits \cite{dillon2017tensorflow, bingham2019pyro}.

This review focuses on the uses of probabilistic models in healthcare. 
At the end of this review, the goal is for the reader to understand how probability plays a role in building models and how it can help address challenges that occur in machine learning models in healthcare. 
This review is organized as follows. The first section sets up mathematical
notation for the basic concepts around probabilistic models. The subsequent 
sections are organized by use cases in healthcare 
that make use of the probabilistic perspective.

\section{PROBABILISTIC MODELS}

To make the distinction between a probabilistic model and a deterministic model clear, we present an example. Consider features $\mbx$ and a count-valued response $y$ like lymphocyte count.
Now take a model with parameter $\mbtheta$ denoted $g_{\mbtheta}(\mbx)$; this model
is a function that predicts lymphocyte count based on the features and
can be learned by finding the parameters that minimize the squared error between the model predictions and the observed response. For simplicity we do not consider a fixed data set, but rather assume access to the data generating distribution $F$. See margin notes for common notation used in this review, including response, probability, and expectation.
With this setup, the model 
$g_{\mbtheta}(\mbx)$ can be trained by finding $\mbtheta$ to minimize
\begin{align*}
	\E_{\mbx, y \sim F}[(g_\mbtheta(\mbx) - y)^2].
\end{align*}
If trained well, the model $g_\mbtheta$ will be close to the expected value of $y$ given
the observed features $\mbx$. This type of model is deterministic and can be used to make predictions. However, without assumptions, this model says nothing about the \emph{distribution} of the response. 

A probabilistic model represents a probability distribution. A probabilistic 
model of a response $y$ given features $\mbx$, rather than being a function, would be a probability distribution $p_\mbtheta(y \g \mbx)$. One way to train such a
model from data is to maximize the likelihood of the observations, 
\begin{align*}
	\E_{\mbx, y \sim F} [\log p_{\mbtheta}(y \g \mbx)].
\end{align*}
If trained well, the model $p_{\mbtheta}$ would be close to the conditional distribution of $y$ given the features $\mbx$ in the data generating distribution. \hspace{-1em}
\begin{marginnote}[]
    \entry{Response/label}{Dependent variable, e.g. patient mortality, denoted $y$}
\entry{Feature}{Independent variable, e.g. patient biomarkers, denoted $\mbx$}
\entry{Parameter}{Model parameters, e.g. coefficients of a logistic regression, denoted $\mbtheta$}
\entry{Expectation}{Expectation with random variables drawn from the distribution $F$, denoted $\E_{\mbx, y \sim F}$}
\end{marginnote}
The probabilistic model $p_\mbtheta$ can not only be used to compute the average value of $y$
given a particular observed feature (by computing $\E_{p_\mbtheta(y \g \mbx)}[y]$),
but also the conditional variance and other statistics as well. Note that binary classifiers are special in that their probability distribution is completely characterized by its expectation.\footnote{In general, binary classifiers can be used to approximate more complex distributions.}
 
Probabilistic models go beyond regression. We list different
flavors of probabilistic models---\textit{predictive models}, \textit{generative models}, and \textit{latent variable models}---in the margin notes. Probabilistic models can refer to both models with probabilistic outputs or probabilistic modeling components in a broader estimation pipeline.
In the subsequent sections, we will go into more detail on the role of 
probabilistic models in healthcare for building predictive models (\Cref{sec:predictive}), for phentopying (\Cref{sec:pheno}), for simulation (\Cref{sec:generative}), and for sequential decision making (\Cref{sec:rl}).

\begin{marginnote}[]
    \entry{Predictive model}{A model for a response given observed features, denoted $p(y \g \mbx$)}
\entry{Latent variable model}{A model that connects unseen traits to observed data, denoted $p(\mbz, \mbx$)}
\entry{Generative model}{A model that outputs samples, denoted $p(\mbx$)}
\end{marginnote}

\section{CHALLENGES IN BUILDING PREDICTIVE MODELS FOR MEDICINE}
\label{sec:predictive}

In this section, we show how probabilistic models can aid in different parts of the model development and model maintenance pipeline in healthcare. Here, we assume that the problem has already been reduced to a collection of features and a (potentially real-valued) response. The goal is to produce a model that predicts the response from the features. The section highlights probabilistic methods---referring to both models with probabilistic outputs or models with probabilistic computation in the development pipeline--to address key challenges in building predictive models for medicine.

\subsection{Missing Values}
\label{sec:missingness}

Missing values are a prevalent problem in clinical data that can impede predictive models, and probabilistic models allow for the modeling of the underlying data mechanism for \textit{missingness}. Healthcare datasets are generally observational and frequently incomplete as a result. Consider a longitudinal clinical dataset of patient visits following a diagnosis of diabetes. For each patient visit, a clinician may choose not to measure all possible biomarker values, resulting in missing values. Patients may also vary in their number of clinical visits, resulting in missing visits for some patients compared to the maximum number of patient visits. Traditionally, machine learning models require completely observed datasets. 
Because removal of missing values may result in a dataset that is too small, or the removal may induce statistical bias, the search for other methods to accommodate missingness is an active area of research.

\begin{marginnote}[]
    \entry{Missingness}{The manner in which data is missing from a sample of the population}
\end{marginnote}

In cases where predictive performance is of greatest importance, the model can directly incorporate missingness. One example might be passing indicators of observation as features, which provide the most information about the response~\cite{miscouridou2018deep}. Using these indicators of observation in a time series, recurrent neural networks have been used to predict patient outcomes in the intensive care unit~\cite{che2018recurrent}. 
Additionally, deep probabilistic models can marginalize missing values to predict time to coronary heart disease~\cite{ranganath2016deep}.

The predictive performance may not be the only quantity of interest for a model. Researchers may also be interested in parameter estimation, e.g., using coefficients to model features importance. A main method to address missing data for parameter estimation is \textit{imputation}, meaning the replacement of missing values based on information from the observed values. After imputation, the transformed data is used for the resulting predictive model. Imputation methods range from using the mean or median of the observed values to prediction of the missing value for each observation. One popular imputation method is multiple imputation using chained equations where missing features are imputed using the posterior predictive distribution of the missing data conditional on the observed data~\cite{white2011multiple}.

\begin{marginnote}[]
    \entry{Imputation}{The replacement of missing values based on information from the observed values}
\end{marginnote}

Implicit in imputation methods is an assumption about the underlying data generating process~\cite{little2019statistical}, namely that the data is either missing completely at random (MCAR) or missing at random (MAR). MCAR refers an assumption that the missingness of a data is completely random and uncorrelated whereas MAR refers to missingness of data that depends on the observed data. Notably, imputation methods cannot support data that is missing not at random (MNAR), meaning the missingness correlates with an unobserved characteristic. Using either the assumption of MCAR or MAR, imputation methods range can leverage probabilistic methods such as Gaussian methods~\cite{zhao2019missing}, causal diagrams~\cite{mohan2018graphical}, or models using auxiliary information~\cite{sadinle2019sequentially}. Identification of MNAR requires additional assumptions, for example semi-parametric estimation using an instrumental variable~\cite{sun2018semiparametric}. Additionally, modeling the data missingness process allows for the model stability when the mechanism for data missingness changes, e.g. across hospitals or across time.

\subsection{Censoring} 
Similar to how probabilistic models can address missing features as described in Section~\ref{sec:missingness}, probabilistic models can capture the probability distribution of the possible outcome events when the patient labels are not observed. 
Labels in healthcare often depend on a patient's
state at some point in the future.  This gap in time between observed features and observed label
means that the labels may be unobserved or \textit{censored} for some patients
\cite{10.1093/biomet/64.2.225}. We may be interested in the time to event, e.g., death, for a patient given observed features and may observe previous patients, only some of whom have observed times to event. A simple machine learning method might regress on the time to event only for patients with observed labels, but this simplification generally underestimates the time to event because patients with a longer gap between observed features and time to event are less likely to be observed.
In contrast, probabilistic models to characterize the \textit{survival function} are trained with the label likelihood and can directly address censored observations by computing the probability that the observed label falls
in the censoring interval through integration when observations are censored at random. 
\begin{marginnote}[]
    \entry{Censoring}{The process through which event times hidden}
    \entry{Survival function}{A function providing a probabilistic estimate of no event occurring before a specified time}
\end{marginnote}
The general assumption that makes survival problems tractable is
censoring at random where the censoring and event time are independent
given the observed features.
Under the censoring at random assumption, consider a patient
that has no event until a censoring time $c$ with features $\mbx$, the
likelihood under a distribution $p$ can be computed as $\int_{c}^\infty p(a \g \mbx) \, da$. 
This approach 
has been used in combination with deep neural networks in recent work on deep survival analysis~\cite{miscouridou2018deep,lee2018deephit,ranganath2016deep,katzman2018deepsurv}.

The evaluation of a survival function requires consideration about the probability distribution of outcomes. The Brier score~\cite{brier1950verification} is the metric traditionally used for estimating the survival function. Evaluation with the Brier score requires appropriate adjustment for censoring by estimating the (inverse) probability of censoring. Non-probabilistic methods like survival forests~\cite{ishwaran2008random} have also been used to incorporate right censoring. Survival forests side-step explicit parametrization of censoring mechanism but build on ensembles of random trees to non-parametrically estimate a cumulative hazard function (a probabilistic quantity) using the Nelson–Aalen estimator~\cite{nelson1969hazard}. 
In the presence of more complex forms of censoring, like interval censoring which is common in epidemiological studies, uncertainty over the interval of censoring should be modeled to fix bias in survival estimates. This statistical bias is particularly problematic if the interval periods themselves are long. Modeling the uncertainties over the censoring mechanism have been demonstrated to improve estimation over imputation techniques~\cite{leung1997censoring} and has been only recently explored in machine learning literature~\cite{cho2019interval} in a non-probabilistic framework using random forests.

\subsection{Calibration}\label{sec:calibration}

Probabilistic machine learning models can ensure that risk scores used in clinical settings are \textit{calibrated}, meaning that the risk estimates accurately characterize the actual risk. Risk scores like the Framingham risk score~\cite{wilson1998prediction} for cardiovascular disease prediction are routinely used for clinical decision making, diagnostic tools, or determining subsequent treatment pathways. There is a general expectation that in addition to predicting the correct binary label $y$ (whether a patient will develop coronary heart disease in $10$ years), the actual risk estimate of the event is available as well. These risk estimates can be obtained only from a machine learning model that frames the supervised learning problem as that of estimating the probability $p_{\theta}(y \g \mbx)$. Therefore, support vector machines (SVMs) will not directly provide such risk estimates without further processing. 

\begin{marginnote}[]
\entry{Calibration}{The measure of how well risk estimates reflect true risk}
\end{marginnote}

\begin{wrapfigure}{r}{0.5\textwidth}
\centering
\includegraphics[width=0.4\textwidth]{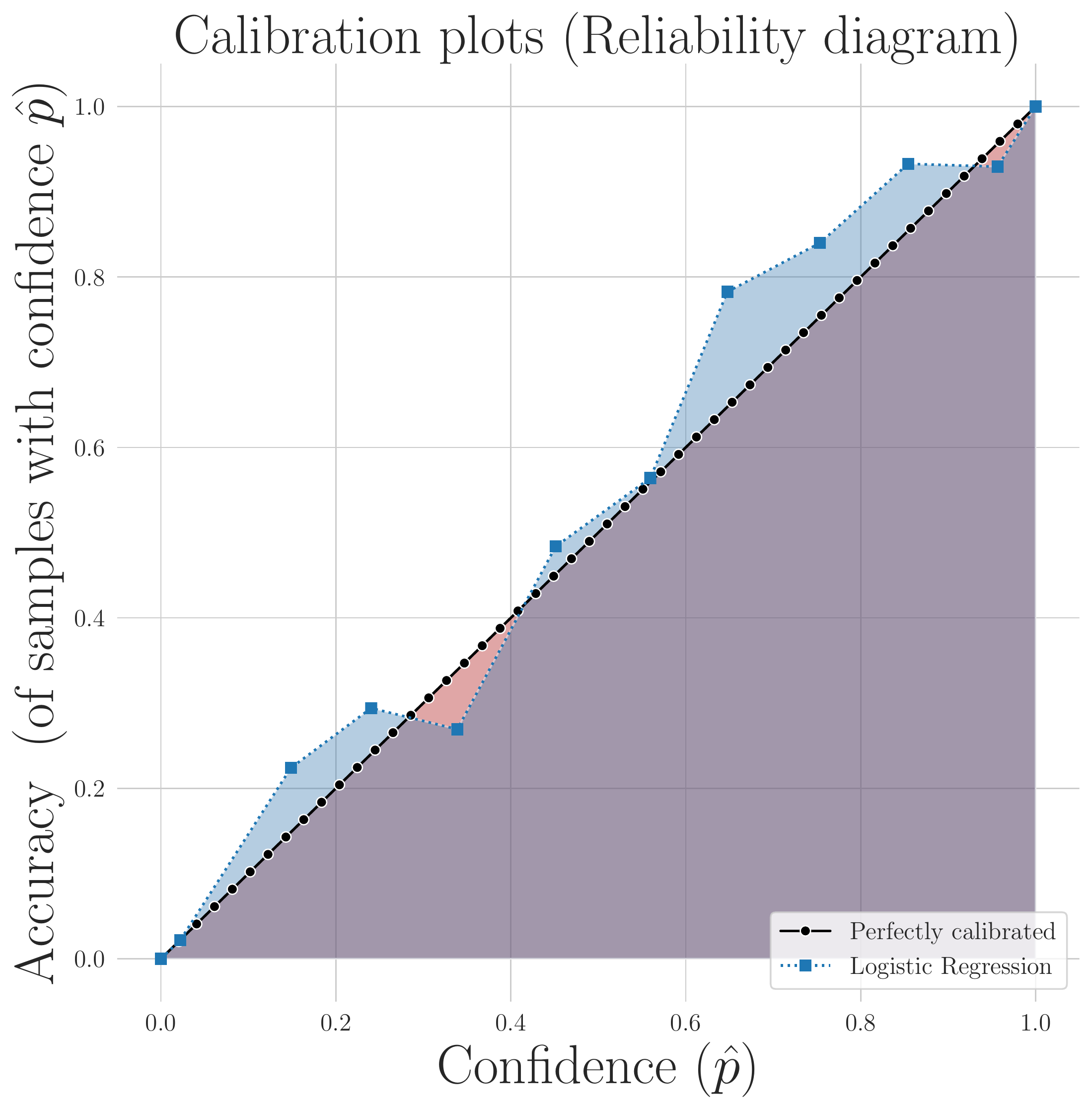}
\caption{Calibration of an machine learning model can be measured via probabilistic modeling. A model is well-calibrated if for all examples, $\mbc$, for which the model provides the same estimate $\hat{p}$ of $p(y \g \mbx)$, the proportion of these examples that are associated with the prediction ($y=1$) is equal to $\hat{p}$. In this figure the red shaded region denotes the reliability plot of a perfectly calibrated model and the blue shaded region from a logistic regression fitted for a synthetic binary classification problem.}
\label{fig:calibration}
\end{wrapfigure}

How well do machine learning models trained to estimate $p_{\theta}(y \g \mbx)$ characterize this risk? This can be understood by quantifying how well-calibrated a model is. An machine learning method is well-calibrated if for all examples, $\mbx$, for which the model provides the same estimate $\hat{p}$ of $p_{\theta}(y \g \mbx)$, the proportion of these examples that are associated with the prediction ($y=1$) is equal to $\hat{p}$. 
Calibration is therefore an inherently probabilistic concept. While traditional machine learning methods like logistic-regression and shallow neural networks typically produce well-calibrated risk scores, modern neural networks notoriously may not~\cite{guo2017calibration}. Further, even if a model is calibrated at the population level, subpopulation level miscalibration can further amplify inequities in clinical decision making~\cite{pfohl2019counterfactual}. Since calibration significantly affects optimality of downstream clinical decision making~\cite{van2019calibration}, diagnostic, and other decision support models should consider probabilistic frameworks for supervised model design.

Calibration of machine learning models can be quantified using reliability diagrams which evaluates the estimated risk of a group of example $\hat{p}$ against the expected sample accuracy for all $\hat{p}$. A well-calibrated model will be close to the identity function in the diagram (see \textbf{Figure~\ref{fig:calibration}}). Assessing the expected (or maximum) difference between the estimated confidence/risk of a model and empirical estimates, a quantity known as the Expected (or Maximum) Calibration Error~\cite{naeini2015obtaining} can summarize how well-calibrated a model is (lower is better). Reliability diagrams can also be assessed at subpopulation levels to determine miscalibration challenges due to lack of samples within groups. Calibration of machine learning methods can be improved post-hoc using techniques such as Platt's scaling~\citep{Platt99probabilisticoutputs}, which corrects for calibration errors after model training by learning a mapping from learned risk scores to calibrated risk scores by estimating a sigmoid mapping on a validation set. These methods can also be used for non-probabilistic models like SVMs that do not produce risk scores as a by-product of supervised learning. Platt's scaling is essentially refitting a logistic regression (a probabilistic model) to obtain risk scores from deterministic models like SVMs. When data is scarce, flexible non-parametric methods like isotonic regression can be used for calibration~\cite{chakravarti1989isotonic}. Sigmoid function refitting as done in Platt's scaling can also be extended to multiclass settings by modeling multiclass problems as a  one-vs-all classification~\cite{zadrozny2002transforming,zadrozny2002reducing}. Calibration can also be improved with other post-hoc methods. Binning methods to obtain calibrated neural networks have also been proposed recently~\cite{kumar2019verified,zadrozny2002transforming}. This class of methods suitably re-estimate risk scores to directly improve/optimize calibration metrics.

\subsection{Uncertainty}\label{sec:uncertainty}
Several types of uncertainties arise when modeling clinical outcomes of interest from finite amount of data. Machine learning model predictions are based on a finite random samples, making the model predictions themselves random. That is, any machine learning model (deterministic or probabilistic) is a function of the random samples from the data generating distribution and hence, has an associated uncertainty. 
The overall uncertainty of a machine learning model prediction, captured by $p(y\g \mbx)$ is known as the \emph{predictive uncertainty} of the model. 
To see how the uncertainty can affect downstream decision making, consider a breast cancer staging model that predicts risk of an adverse event. If the model-estimated disease stage has different variances for different features values, the resulting decisions can be suboptimal if a single decision threshold is used.

Predictive uncertainty can be further decomposed into different sources. 
The first source, called aleatoric uncertainty,
measures the noise in the labels in the true
data generating process.
For instance, diagnostic labels can have some uncertainty when different clinicians annotate the same sample~\cite{raghu2019direct}. 
The second source stems from the uncertainty about an estimated model's match to the true data generating distribution. This uncertainty is called model uncertainty, also known as epistemic uncertainty.
Model uncertainty is a combination of the choice of the model class used to approximate a property of the true data generating distribution, as well as the fact that
multiple parameters $\mbtheta$ within the same model class can approximate the data well. 


When a large enough sample size is available and assuming correctness of the model class, uncertainty can be quantified using asymptotic analyses~\cite{de1981asymptotic}. Large scale datasets may not always be available. In this case, uncertainty is either captured via \textit{bootstrapping} samples~\cite{efron1983leisurely} without explicit probabilistic modeling or 
using a fully \textit{Bayesian framework}. 
\begin{marginnote}[]
\entry{Predictive uncertainty}{An expression of the statistical dispersion of the model prediction}
\entry{Bootstrapping}{A procedure involving sampling with replacement, which can be used to quantify uncertainty over a data sample}
\entry{Bayesian framework}{A class of statistical methods that assign probabilities or distributions to events or parameters based on prior knowledge before experimentation}
\end{marginnote}
Bootstrapping is a sampling with replacement procedure, a proxy to quantify uncertainty over the data sample, so that multiple model estimates can be obtained. Predictions from different candidate models then give an estimate of variability in predictions, i.e., capturing uncertainty. Neural networks can also be retrained on bootstrapped samples to quantify this uncertainty although performance may degrade with fewer data samples~\cite{lakshminarayanan2017simple}. 
These methods can be used to assess overall uncertainty as well as calibration of a model, 
but still only provide expectations over outcomes rather than a full characterization of the predictive uncertainty. Thus, the onus is on the end user to calculate probabilistic quantities of interest like variance that will be useful for downstream decision making.

In a full Bayesian framework, the outcome of interest is naturally modeled as a distribution i.e. $p(y\g \mbx)$. Additionally, distributional assumptions are made over the model parameter $\mbtheta$ itself allowing to fully characterize as well as decompose the predictive distribution. The relationship between overall uncertainty and model uncertainty $p(\mbtheta \g \mathcal{D})$ (where $\mathcal{D}$ is the dataset sample from the generating distribution) is captured by the following relationship:
\begin{align*}
    p(y\g \mbx) = \int_{\mbtheta} p(y \g \mbx, \mbtheta) p(\mbtheta \g \mathcal{D}) d\mbtheta
\end{align*}

Notably, such uncertainty estimation via Bayesian modeling is more challenging to incorporate for non-probabilistic machine learning models like support vector machines. For deep neural networks, for capturing this uncertainty over predictions, i.e. $p(y\g \mbx)$, Bayesian deep learning~\cite{gal2016uncertainty} is commonly employed. These methods usually capture intermediate uncertainties i.e. over parameters $p(\mbtheta \g \mbx)$ during model training. For instance, one method~\cite{gal2016dropout} reinterprets traditional regularization methods like dropout regularization~\cite{gal2016dropout} in a probabilistic framework. However, eliciting good prior distributions over parameters, required for Bayesian modeling, can be a challenge for modern deep neural networks. Even in simple models, misspecification of such priors can lead to unreliable uncertainty estimates~\cite{grunwald2017inconsistency}. 
Recently, extensive evaluation of Bayesian neural networks and neural network ensemble methods for uncertainty characterization has been demonstrated on critical care datasets eICU and MIMIC-III~\cite{dusenberry2020analyzing}. Similar to bootstrapping neural network ensembles do not explicitly model the predictive distribution to obtain uncertainty estimates. In one example, model predictions and hence downstream decisions can vary widely for individuals due to these uncertainties and quantifying this uncertainty can reduce the possibility of sub-optimal clinical decisions~\cite{dusenberry2020analyzing}.

\subsection{Data Shift}

\begin{figure}[t]
  \subfloat{
	\begin{minipage}[c][1\width]{
	   0.3\textwidth}
	   \centering
	   \includegraphics[width=1\textwidth]{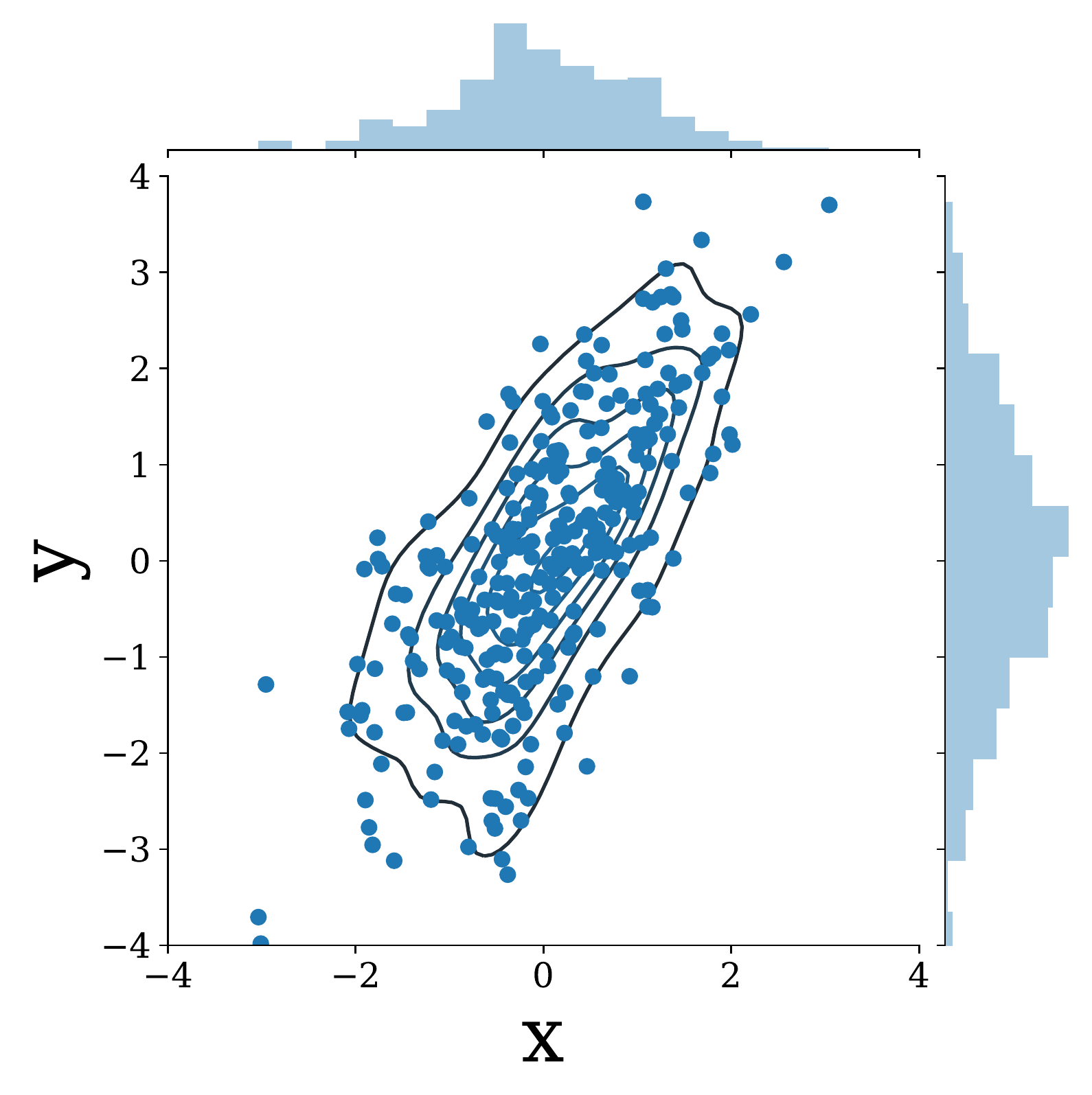}
	\end{minipage}}
 \hfill 	
  \subfloat{
	\begin{minipage}[c][1\width]{
	   0.3\textwidth}
	   \centering
	   \includegraphics[width=1\textwidth]{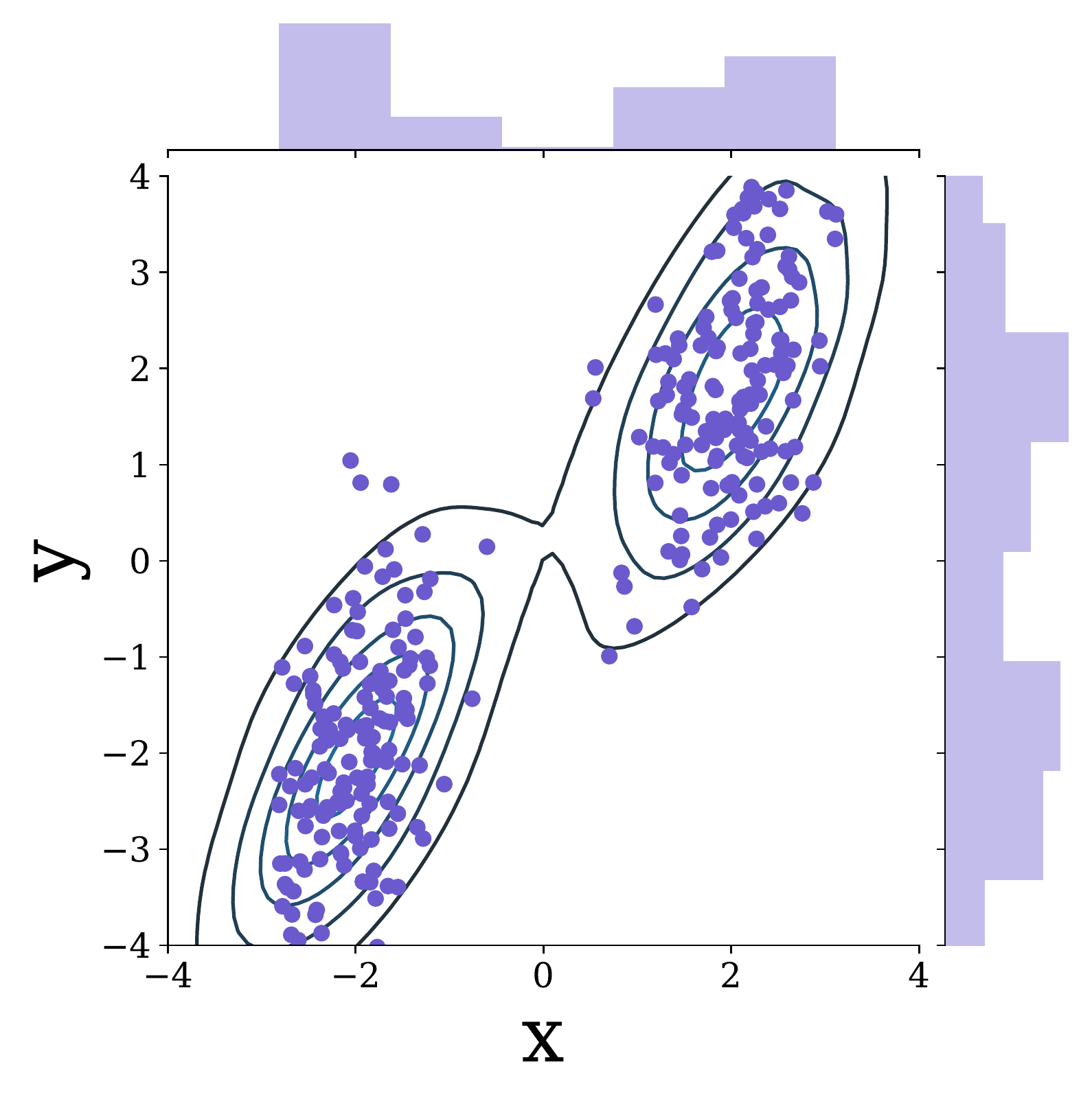}
	\end{minipage}}
 \hfill	
  \subfloat{
	\begin{minipage}[c][1\width]{
	   0.3\textwidth}
	   \centering
	   \includegraphics[width=1\textwidth]{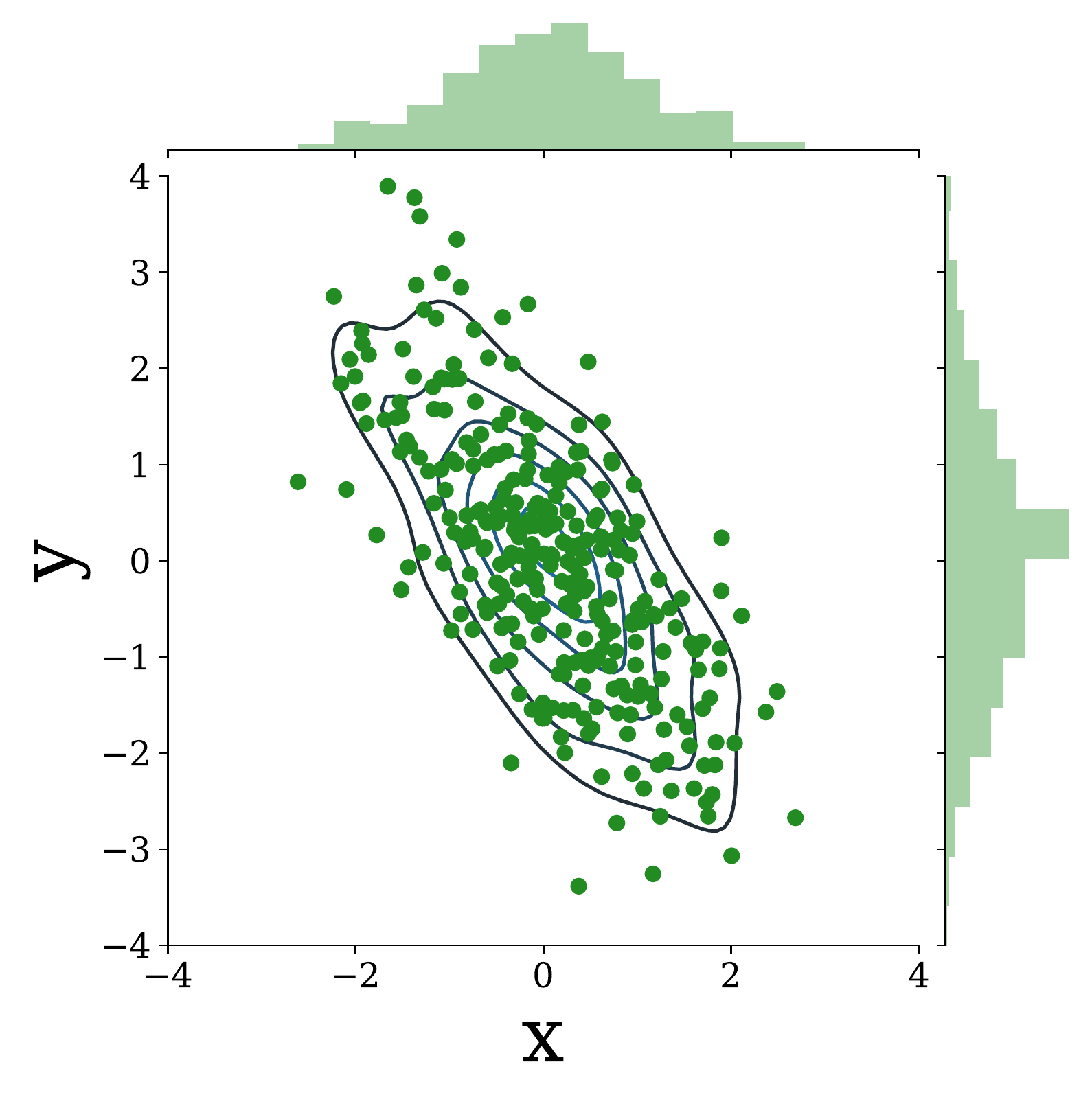}
	\end{minipage}}
\caption{Consider a data distribution of $p(x,y)$ \textbf{(left)} where $x$ and $y$ are continuous and univariate and $y = x + \epsilon$ with $\epsilon \sim \mathcal{N}(0,1)$. Covariate shift \textbf{(middle)} refers to changes in $p(x)$, for example because of differences in patient demographics between hospitals. Here $p(x)$ changes from a unimodal distribution to a bimodal distribution. Label shift \textbf{(right)} refers to changes in the relationship between $x$ and $y$, for example because of differences in treatment policies between hospitals. Here, we show $y = -x + \epsilon$ with $\epsilon \sim \mathcal{N}(0,1)$.}
\label{fig:data_shift}
\end{figure}

After clinical model development, the data setting in which the model is used may differ from the setting in which the model was trained, a shift that can be characterized and accommodated by probabilistic models. 
For example, model performance may degrade when the data distribution for development and for deployment differ across locations~\cite{zech2018variable} or across time~\cite{nestor2019feature}. This shift
of data can threaten the performance of machine learning models in deployment. Consider a setting where a model is trained to predict an outcome $y$ based on patient features $\mbx$. When a model is trained on a source domain with data distribution $p(y\g\mbx)$, the concern is that the setting of deployment may have a different distribution in $p(\mbx)$ (\textit{covariate shift}) or $p(y\g\mbx)$ (\text{label shift}), as seen in \textbf{Figure~\ref{fig:data_shift}}.
Probabilistic models can describe this process and improve robustness of models over different data distributions.

\begin{marginnote}[]
    \entry{Covariate shift}{The condition where the feature distribution changes}
\entry{Label shift}{The condition where the relationship between features and response changes}
\end{marginnote}

Probabilistic models can address covariate shift through identification and correction. Covariate shift is the condition where the feature distribution changes. 
For example, urban and rural hospitals have differences in clinical care patient populations. Rural hospitals have higher prevalence of chronic obstructive pulmonary disease including higher rates of Medicare hospitalizations and death~\cite{croft2018urban}. 
In model settings with long time-series sequences, for example continuous monitoring in medical settings, change point detection identifies abrupt changes in observed data where $p(\mbx)$ shifts dramatically~\cite{yang2006adaptive, malladi2013online}. In settings with clear and known demarcations between source and target domains, for example across hospitals, probabilistic models can adapt model development on the source domain to a target domain for deployment. One technique is using importance sampling, based on estimates of the probability of the domain given the features, to reweight observed datapoints~\cite{shimodaira2000improving,sugiyama2006importance}. Modeling and correcting for data shifts allows for more accurate and robust clinical models.

Label shift, meaning $p(y\g\mbx)$ changes, is the mechanism of labels conditional on features changes and may be due to changes in \textit{treatment policy} or induced by the clinical model itself~\cite{subbaswamy2020development}. For example, clinical protocol for disease management may change as additional scientific discoveries shape medical knowledge, resulting in multiple myeloma survival rates improving dramatically with one-year relative survival rates increasing from 69\% in 1973 to to 82\% in 2013 due to better treatment policies~\cite{thorsteinsdottir2018dramatically}. Without reasoning about this shift, models may yield erroneous and outdated results. In one famous example, a model predicting pneumonia risk stratification did not account for patients with asthma being given more severe interventions; the predictions for pneumonia patients with asthma yielded lower risk when clinical literature reveals the opposite~\cite{caruana2015intelligible}. In that case, models can be trained to be anticipate shifts in policy based on the treatment policy distribution~\cite{schulam2017reliable} and therefore more readily update for changes in clinical treatment protocol. 

\begin{marginnote}[]
\entry{Treatment policy}{A medical protocol for interventions on a given patient}
\end{marginnote}

Alternatively, the model itself may induce change in the data after deployment.
For example, a diagnostic model may yield predictions that guide treatment policy where high risk patients receive more aggressive treatments, meaning the model may not be calibrated to the outcomes that occur after model deployment. One method to address this shift is to produce probabilistic models that yield stable and calibrated predictions for not only pre-deployment outcomes but also against future outcomes that manifest from the data shifts~\cite{perdomo2020performative}. Strategic behavior from the individuals may also affect the data distribution, for example, with otherwise healthy patients adapting to the deployed clinical model for treatment and displaying behaviors similar to sick patients in order to receive additional treatment. Causal inference methods can model complex systems to address changes in strategic behavior~\cite{bottou2013counterfactual,milli2019social}. 

\subsection{Fairness}

The fairness of a clinical model is a key concern to ensure unbiased predictions for medical tasks, and probabilistic models help enable the defining, auditing, and ameliorating this algorithmic bias. Because of the high-stakes setting of healthcare, researchers have raised numerous concerns about fairness and algorithmic bias in clinical contexts~\cite{adamson2018machine,obermeyer2019dissecting,rajkomar2018ensuring}. There exist many definitions of algorithmic fairness~\cite{hardt2016equality,dwork2012fairness,zemel2013learning,bolukbasi2016man}, and probabilistic models are crucial to representing and addressing many notions of ethics and disparity.

One branch of fairness focuses on the notion of \textit{individual fairness}~\cite{dwork2012fairness}. This fairness definition implies that a patient with similar characteristics to another patient should not receive worse clinical care than the other. 
A key requirement is that similar patients receive similar probabilistic predictions, using some notion of similarity~\cite{bolukbasi2016man}.

Another branch of concerns analyzes \textit{fairness across groups}. In this family of fairness definitions, algorithmic bias is assessed based on known and pre-defined sensitive attribute groups, e.g. race, gender, socioeconomic status. Because some definitions are impossible to satisfy simultaneously~\cite{chouldechova2017fair,kleinberg2016inherent}, definition choice is crucial in different health settings. 

The fundamental idea across group fairness definitions is that a definition of performance should not differ across groups and statistically significant violations of this assumption could prove algorithmic bias. 
 \begin{marginnote}[]
\entry{Algorithmic fairness}{The study of definitions related to the justice of model predictions}
\entry{Individual fairness}{The principle that similar individuals should be treated similarly}
 \entry{Fairness across groups}{The principle that pre-defined patient groups should receive similar model performance}
\end{marginnote}

Although not all definitions of group fairness require probabilistic models, we outline a few examples of group fairness definitions that leverage the data distributions. 
For example, the two groups may not adjust to changes in data shift in the same way. Probabilistic models can learn models on the source data distribution that will adapt to a target data distribution while satisfying group fairness definitions~\cite{singh2019fair}. In another example, examining the calibration of an algorithm across each group may reveal larger disparities~\cite{pleiss2017fairness,simoiu2017problem}, and health settings may require calibration across multiple subgroups~\cite{hebert2018multicalibration}. 
Lastly, probabilistic models can reason about regression on continuous outcome variables, for example in the case of health care costs. When comparing predictions for healthcare costs between patients with mental health and substance abuse disorders compared to patients without them, the sensitive attribute group and the residual error from a learned model should be independent. Researchers can then measure the covariance between the sensitive attribute group and the residual error as a proxy for independence to detect the level of algorithmic bias in the model~\cite{zink2019fair}.

The quest to propose solutions to algorithmic bias is still in its infancy. Notably, any balance between algorithmic fairness and accuracy may be ethically challenging for health settings. Although non-probabilistic models solutions to algorithmic bias certainly exist~\cite{obermeyer2019dissecting}, probabilistic models consider how to fix discrimination within the constraints of the model. 
One method seeks to induce independence between model predictions and the sensitive attribute through a latent representation can be learned that maximizes performance and minimizes dependence on the sensitive attribute~\cite{zemel2013learning}. Another approach alters the objective function of the predictive model to regularize for algorithmic fairness based on a specified definitions~\cite{bechavod2017learning, zafar2017fairness}. Lastly, probabilistic models can reason about the effectiveness of additional data collection or other actions using the existing data~\cite{chen2018my}.

\section{PHENOTYPING WITH LATENT VARIABLES}
\label{sec:pheno}

We now pivot to three major healthcare application areas where probabilistic modeling has been extensively used, starting with phenotyping. Phenotyping~\cite{richesson2013electronic} is the process of producing concise representations of medical concepts or diagnoses composed of observable clinical traits that could be used to facilitate cohort selection or trait definition~\cite{banda2018advances}. 
From a latent variable perspective, we can hypothesize that simple latent ``summaries'' explain the variation in observed patient records, and serve as a governing factor in determining how different patients will progress~\cite{alaa2019attentive} or react to different interventions~\cite{ghassemi2016understanding}. Prior work has identified many forms of electronic phenotyping, including rule-based methods, text processing, noisy data learning, and unsupervised discovery of latent \textit{phenotypes}~\cite{advances2018shah}.  
We categorize these efforts into three settings, based primarily on the amount of the level of label supervision used.
\begin{marginnote}[]
    \entry{Phenotype}{The presentation of characteristics of an individual}
\end{marginnote}

\begin{figure}[ht]
  \subfloat{
	\begin{minipage}[c][1\width]{
	   0.45\textwidth}
	   \centering
	   \includegraphics[width=1\textwidth]{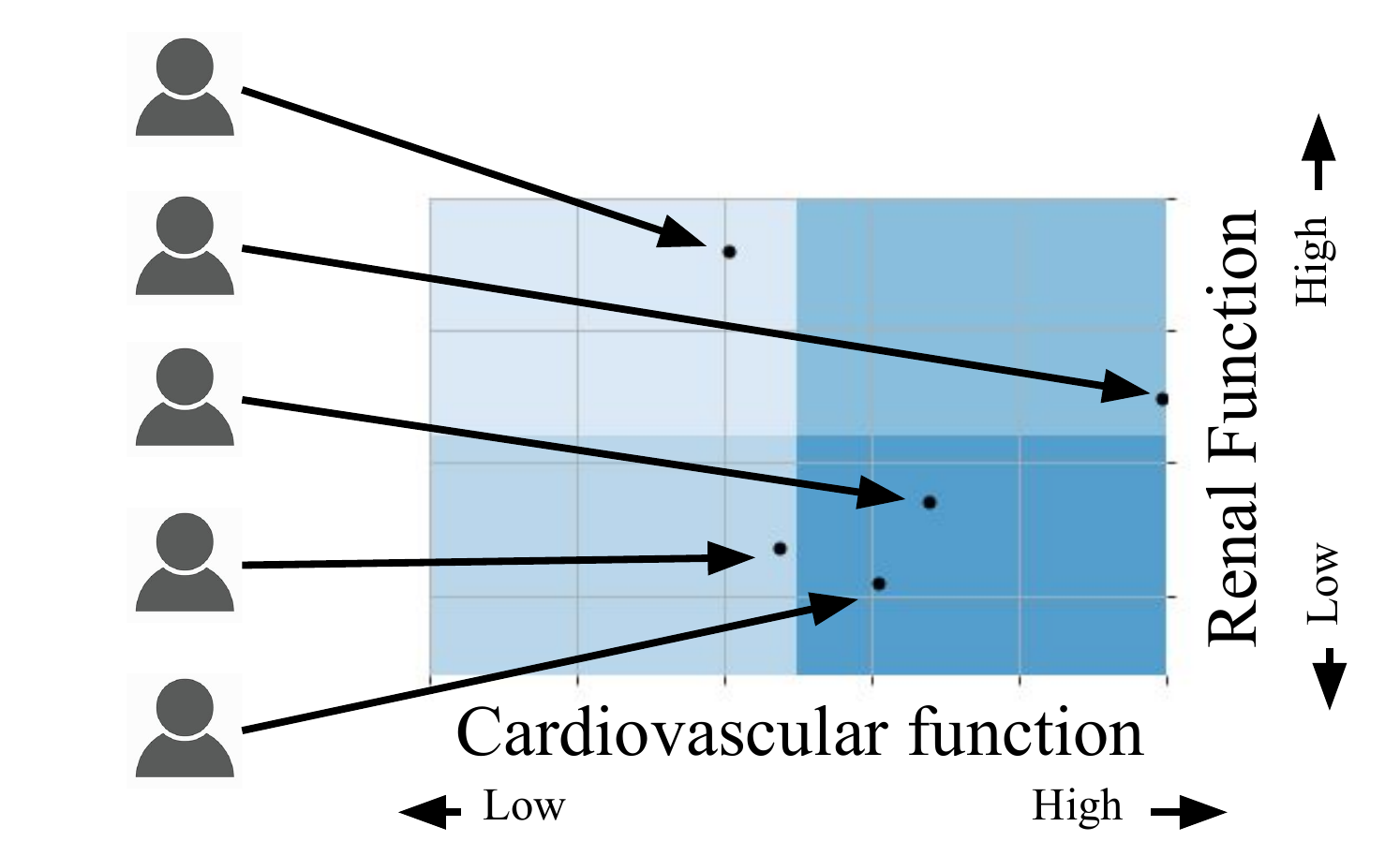}
	\end{minipage}}
\hfill 	
  \subfloat{
	\begin{minipage}[c][1\width]{
	   0.45\textwidth}
	   \centering
	   \includegraphics[width=1\textwidth]{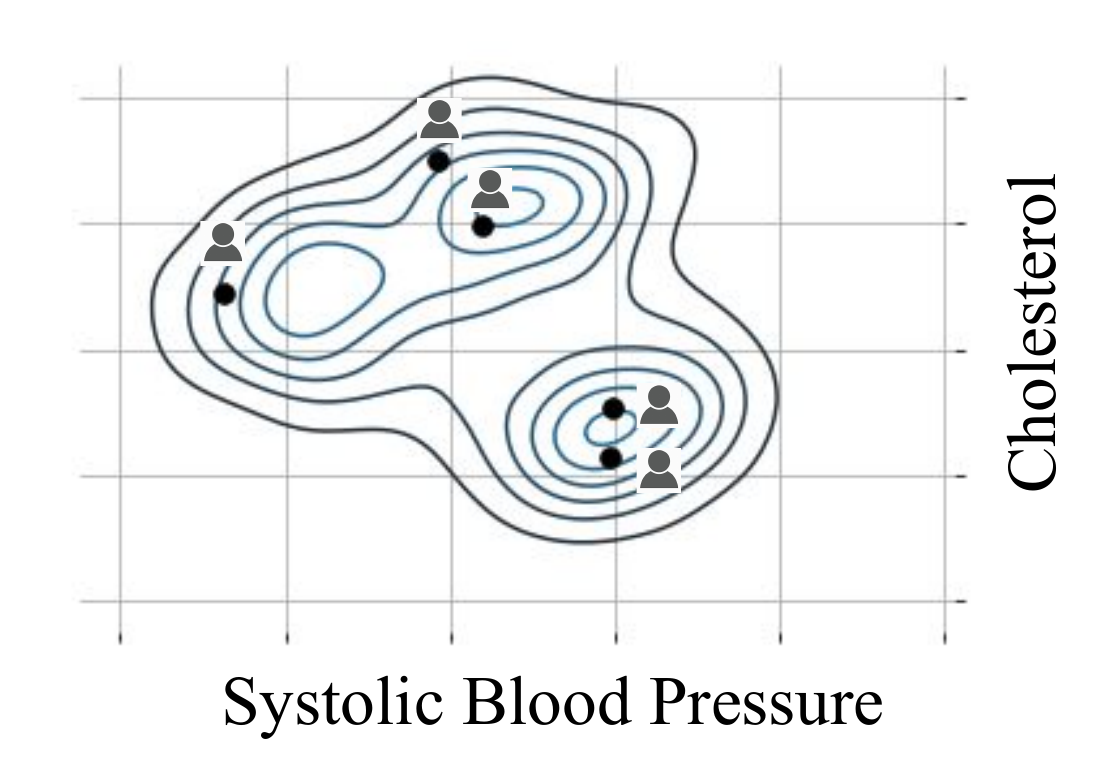}
	\end{minipage}}
\caption{Phenotyping matching~\textbf{(left)} matches patients $x$ with explicitly defined phenotypes, here described in different shades of blue as having high/low cardiovascular function and high/low renal function. Uncovering latent phenotypes~\textbf{(right)} learns new phenotypes to identify useful characterizations through probabilistic clustering.
}
\label{fig:phenotyping}
\end{figure}

In some settings the goal is \emph{phenotypic matching}, where phenotypes are explicitly defined, and the goal is to map noisy data sources into these labels. In others the goal is \emph{uncovering latent phenotypes}, where there is uncertainty about what phenotypic definitions should be, and the goal is to identify useful characterizations that could impact patient care~\cite{doshi2014comorbidity}. \emph{Semi-supervised phenotyping} is a hybrid approach that straddles matching and discovery, where we assume that \emph{some} label information is available, e.g., with the use of specific ``anchoring'' clinical terms \cite{halpern2016electronic,yu2018enabling}. Both phenotypic matching and uncovering latent phenotypes are visualized in \textbf{Figure~\ref{fig:phenotyping}}.
In this review, we review each of these three settings---phenotype matching, uncovering latent phenotypes, and semi-supervised phenotyping---outlining key areas where probabilistic models have made, or could make, an impact. 

\subsection{Phenotypic Matching}
Phenotypic matching is a computational approach that involves matching or summarizing patient records $\mbx$ into validated phenotypes $\mbz$. This style of phenotyping is useful when a meaningful classification or alert could be automated for patients that need to be treated early, or more efficiently. For example, structured electronic health record (EHR) data in the form of International Classification of Disease (ICD) billing codes and medication prescriptions have been used to detect coronary heart disease or rheumatoid arthritis \cite{kirby2016phekb}. More recent work has also targeted use of natural language processing \cite{liao2019high}, with additional unsupervised learning of important features from the structured EHR~\cite{zhang2019high}. Importantly, phenotypic matching assumes that the existing phenotype definitions $\mbz$ are appropriate, robust and identifiable from the given data $\mbx$. Known challenges in label leakage, soft labels, confounding and missingness must therefore be considered very carefully \cite{ghassemi2019practical}.  

Probabilistic methods could be particularly useful in phenotyping due to the inherent uncertainty in many phenotype definitions, which can change over time as in autism~\cite{greenspan2018autism}. Some work has moved to integrate notions of uncertainty into matching, either by performing large-scale phenotype estimation from observational data~\cite{pivovarov2015learning}, or by visualizing probabilistic estimates over phenotypes for interactive verification~\cite{hirsch2015harvest}. More generally, models that attempt to discover phenotypes based on underlying data regularities, or use prior labels in a semi-supervised fashion, are well-suited for probabilities modeling. 

\subsection{Uncovering Hidden Phenotypes}
Instead of matching to a known phenotype $\mbz$, many problems in healthcare require identifying potential phenotypes from data. 
When phenotypes are unknown, they are hidden. This makes finding unknown phenotypes well-suited to latent variable models.
This style of model takes the phenotype of a person to be a latent
variable $\mbz$ with a hypothesized prior distribution $p(\mbz)$. The latent variable controls the distribution of the observed data $\mbx$ through the likelihood $p(\mbx \g \mbz)$. As a whole, the model with parameters $\mbtheta$ is $p_\mbtheta(\mbz, \mbx)$. 
The goal during learning is to find parameters of the latent variable model that make the observed data likely. Typically, this is done through an approximation to the posterior $p(\mbz \g \mbx)$, which also helps compute phenotypes given an observation from a single patient.

To illustrate, imagine that $\mbx$ is a vector of measured traits like hemoglobin A1C---a test that evaluates the average glucose in blood over the past few months, blood pressure---a measure of the pressure that blood exerts within arteries, and cholesterol---a waxy substance that can develop into fatty deposits in blood vessels. Let the phenotype $\mbz$ be a binary variable marking a patient as belonging into one of two possible phenotypes. The likelihood is set up such that the phenotype is encouraged to explain relationships between the observed traits, e.g., blood pressure and cholesterol have a positive correlation. 

One salient example of a latent variable model is a variational autoencoders (VAEs) \cite{kingma2013auto}, where neural networks are both used to parametrize the likelihood and the posterior approximation.
Recent work has used VAEs in a health setting to create low-dimensional latent representations of a phenotype feature space that additionally capture individual rates of change along each dimension during aging \cite{pierson2019inferring}. We note that non-probabilistic methods have also identified meaningful subtypes including asthma subtypes through hierarchical clustering~\cite{doshi2014comorbidity} and type 2 diabetes through tensor factorization~\cite{henderson2018phenotyping}. 

Uncovering latent phenotypes (and subcategories) often aims to discover similar patient subgroups that may share the same underlying disease mechanism, or treatment response patterns. In conditions such as asthma, where there is heterogeneity in symptom expression and response to therapy, phenotyping subcategories of $\mbz$ into $z_1 \ldots z_k$ can be particularly useful. Data-driven methods for data-driven characterization of wheeze patterns, and the further discovery of biomarkers to identify phenotypes are useful, particularly in childhood \cite{belgrave2013characterizing,deliu2017asthma}.
Treatments themselves can also be heterogenous, and clustering them can reveal treatment profiles at a large scale that would not have been clear in a smaller sample~\cite{hripcsak2016characterizing}. For example, recent work in endometriosis subtyping provides evidence of different treatments in each of four subtypes learned in an unsupervised mixed-membership model \cite{urteaga2020learning}. Importantly, while the subtypes present new knowledge, they also align well with previous clinical knowledge about endometriosis, and reflect direct patient experiences with endometriosis.

Probabilistic clustering of time series specifically can be high-value in clinical settings, because sequential clinical markers from different patients could be heterogeonous. For instance, autoimmune diseases are known to be heterogenous, and hierarchical probabilistic models can be used to infer disease subtypes in such patients and explain away correlations that are not relevant to the question of interest~\cite{schulam2015clustering}. 
In more acute settings, neural networks and switching state autoregressive models have been used in the intensive care unit to predict upcoming interventions~\cite{ghassemi2017predicting,suresh2017use}. These estimations can be used to group patients that are ``maximally activiating'' for intervention onsets~\cite{suresh2017clinical}. 

\subsection{Semi-supervised Phenotyping with Anchors}
Phenotypic matching can initially be very labor intensive, as it requires many manual gold standard annotations of $\mbz$. In some cases, there is some strong information that can be used for partial phenotype matching, but other facets of the phenotype must be discovered. 
One potential solution to this is to assume that phenotype labels themselves are weak or ``semi-supervised'', with only a few known features from data being clear conditional markers. 
In this setting, probabilistic modeling approaches can be used to identify additional features that correlate with the anchoring clinical terms \cite{halpern2016electronic,yu2018enabling}.
Anchoring can be thought of as a form of noisy labelling, where observing the positive \textit{anchor} $a_i$ unambiguously reveals the state of latent phenotype $z_i$ to be positive, but a negative anchor $a_i$ reveals nothing about $z_i$, so that $p(z_i = 1\g a_i = 1) = 1$. 
\begin{marginnote}[] 
    \entry{Anchor}{A specified feature which, if positive, reveal that the desired attribute is also positive}
\end{marginnote}
Other work has similarly used anchor words to provide a form of supervision in topic modeling for characterizing pancreatitis outcomes \cite{chen2017survival}. 
Further work in phenotype inference with semi-supervised approaches have used mixed membership models \cite{rodriguez2019phenotype} to inferring binary labels for clinical condition targets when trained on limited samples.

\section{GENERATIVE MODELS}
\label{sec:generative}

\begin{wrapfigure}{r}{0.4\textwidth}
    \centering
    \includegraphics[width=0.4\textwidth]{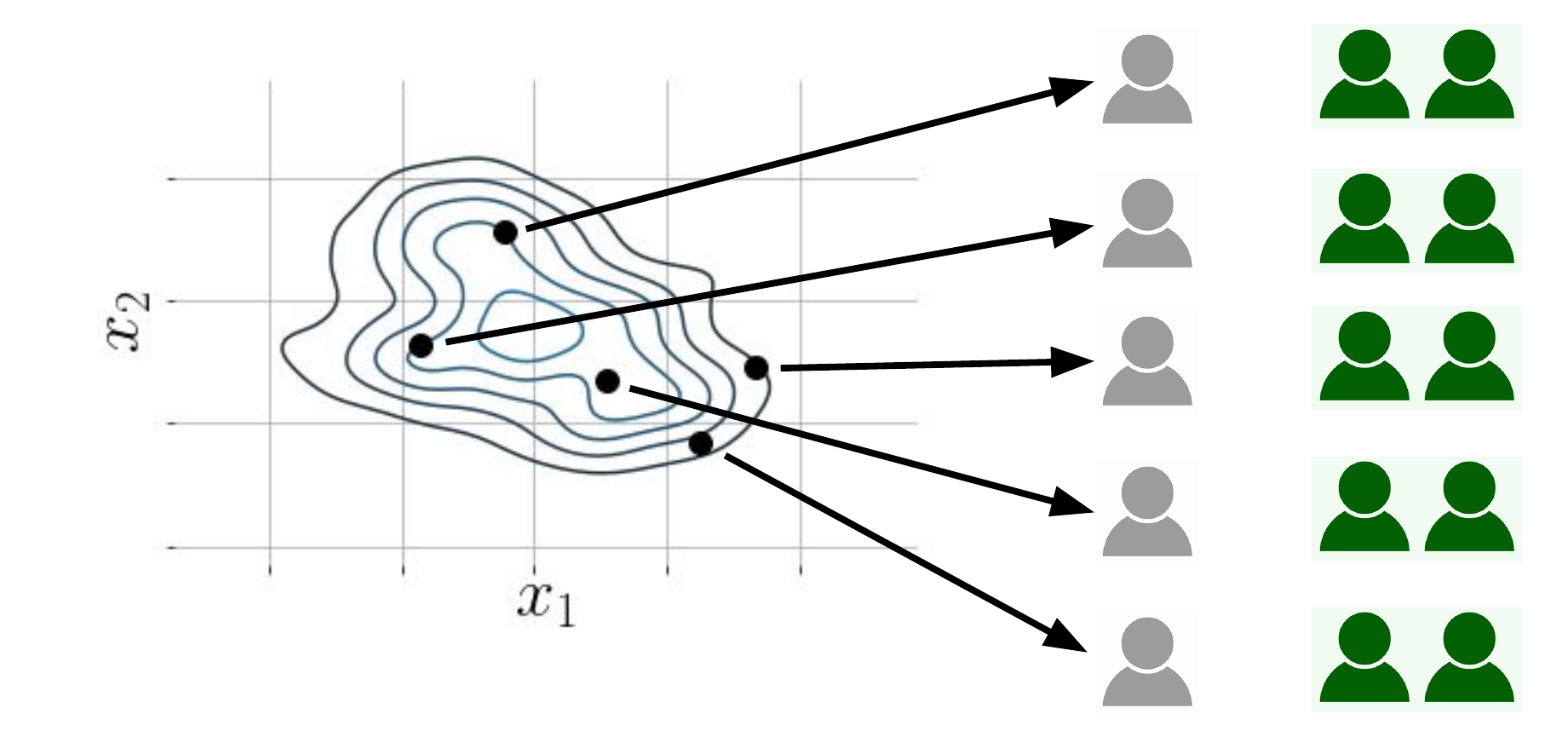}
    \caption{Generative models produce samples from data distribution $p(\mbx)$. Synthetic data samples (grey patients) resemble actual patient data (green patients) and are useful for many clinical problems including data augmentation and abnormality detection.}
    \label{fig:generative}
\end{wrapfigure}

Generative models refer to the class of probabilistic models trained to produce samples that match samples from the distribution from which observed data is collected.
For example, given data vector $\mbx$, we can \textit{sample} from $p(\mbx)$ using a generative model to produce synthetic but realistic data, as shown in \textbf{Figure~\ref{fig:generative}}.
Recent advances in deep learning have led to promising generative models. Generative adversarial networks (GANs)~\cite{goodfellow2014generative} use two neural networks to first create artificial imitations of the training data and then separately decide whether a given sample was genuine or counterfeit. Other models include deep likelihood models like
normalizing flows~\cite{rezende2015variational} where a simple initial density is transformed into a more complex one by applying a sequence of invertible transformations.

\begin{marginnote}[]
\entry{Sampling}{The act of generating data points from a given distribution}
\end{marginnote}

In this review, we focus on three promising cases for generative models in medicine. First, we describe the the use of generative models to overcome data deficiency or accessibility challenges in clinical data. Next, we describe how generative models can be used for clinical tasks like abnormality detection and modality translation in medical imaging. Lastly, we address the generation of viable candidates in the drug discovery process.

\textbf{Data augmentation.} Clinical predictive models often suffer from poor performance because of insufficient data stemming from patients with rare conditions, labels that are costly to obtain, or other logistical challenges. Data augmentation using generative models can create synthetic data that is similar to the true dataset of interest to mitigate the effects of insufficient data. Particularly in medical imaging~\cite{shorten2019survey}, generative models can then accelerate model development that would be otherwise impeded by small dataset sizes.

Data augmentation can address small or imbalanced datasets through the generation of synthetic datapoints. 
For example, GANs have been used to augment all three classes of liver lesions~\cite{frid2018gan}, to increase the number of positive examples of bone lesions~\cite{gupta2019generative} or brain metastases~\cite{han2019learning}, and to provide additional red blood cell images for downstream segmentation tasks~\cite{bailo2019red}. 
Note that when data augmentation is applied to a dataset, the augmented data does add any additional information, but the augmented data can regularize the model or help mitigate class imbalance.

Data augmentation can address privacy and security concerns that impede model development by creating synthetic data that can be shared across institutions. Researchers created synthetic patients using a GAN based on actual patients in a systolic blood pressure trial using GANs with differential privacy constraints. The synthetic patients were similar enough to the original trial patients that models trained on each dataset yielded effectively the same results~\cite{beaulieu2019privacy}. Machine learning models trained using the synthetic data generalized well to the original training data. Because sharing individual-level data across clinical institutions often requires close collaboration and extensive data use agreements, data augmentation and distribution of privacy-preserving synthetic data can circumvent these restrictions.

\textbf{Clinical task assistance.} Generative models can assist with clinical tasks including abnormality detection and modality translation in medical imaging~\cite{kazeminia2018gans}. With regards to abnormality detection, images from healthy patients are used to learn a latent space. Then, a reference medical image potentially containing an abnormality is encoded into the healthy latent space. Samples are drawn from the latent space near the encoded image and compared against the reference medical image. Any differences between the generated samples and the reference medical image are highlighted and analyzed for medical abnormalities~\cite{chen2018deep, baumgartner2018visual}. The task of translation of medical imaging between modalities --- e.g., positron emission tomography (PET) scan to computed tomography (CT) scan --- is necessary for tasks like attenuation correction of PET data using CT scans. Two different encoders are learned for two medical imaging modalities with the same latent space. Then, one image can be translated into another modality by mapping first to the latent space and then decoding to the second modality~\cite{armanious2020medgan}.

\textbf{Drug development.} Currently the development of \textit{de novo} drug-like compounds relies on the identification of new molecular entities. While prior methods have relied on manual selection of candidates or molecular predictors to estimate viability, sampling from generative models can create large virtual chemical libraries, which can then be screened more efficiently for \textit{in silico} drug discovery purposes. 

Methods for candidate generation focus on use cases that are goal-based or distribution-based~\cite{brown2019guacamol}. Goal-based describes methods that generate molecular structures that perform well according to a scoped goal like structural or physiochemical features, and they do not necessarily rely on probabilistic models. Distribution-based methods observe acceptable molecules and generate similar molecular structures, a task well-suited for generative models. Using GANs, researchers have produced novel small-molecule organicestructures~\cite{putin2018adversarial}. However, molecules generated by GANs may lack diversity~\cite{preuer2018frechet}, a problem that other generative models such as VAEs do not suffer~\cite{segler2018generating}. One model combines distribution-based and goal-based
approaches 
by learning a latent representation of molecules using a VAE and then using reinforcement learning using a reward function, similar to the goal-based scores, to explore the space~\cite{zhavoronkov2019deep}.

\section{REINFORCEMENT LEARNING FOR TREATMENT PLANNING}\label{sec:rl}

Sequential decision making is a core part of healthcare~\cite{futoma2020popcorn,shortreed2011informing,chang2019dynamic,prasad2017reinforcement,komorowski2016markov,martin2009reinforcement,li2018actor,yu2019reinforcement}, and probabilistic models enable a variety of approaches including characterizing randomness in patient disease progression and stochasticity of clinician intervention practices~\cite{foxdeep,gottesman2019evaluating,futoma2020popcorn}.
Consider a patient admitted to the intensive care unit (ICU) as they develop a respiratory failure. Clinicians will attempt to manage the patient's condition with a series of advanced respiratory interventions over the coming hours/days with the goal of restoring their function for a successful discharge and survival. During the ICU stay, a sequence of interventions are done. For example, starting with a mechanical ventilation for urgent and initial resuscitation of the patient, to prevent further deterioration. After this is normalized, further interventions like more therapeutic treatments and additional differential diagnosis follow. This is an example of clinical care that is a series of interventions to improve the chances of patient survival. Similarly, treating progression of chronic conditions is a sequential set of interventions to manage disease severity, just over months/years of disease progression. In machine learning, the closest analogue to modeling this problem is known as Reinforcement learning (RL). The primary goal of an RL algorithm can be to either learn a function or a distribution over possible interventions given patient state (policy learning) or evaluating the potential reward of an existing policy (policy evaluation).

RL can be seen as a generalization of supervised machine learning learning where the goal is to make a sequence of optimal decisions to maximize long term \emph{rewards}~\cite{yu2019reinforcement} (patient survival in the ICU example). In this setup, a clinician (learning agent) interacts within an environment (patients) via actions (ventilation and other interventions) it performs, observes the changes in the environment (patient state) in order to successfully discharge the patient (maximize a longer term numerical reward)~\cite{sutton2018reinforcement}.

\begin{marginnote}[]
    \entry{Reinforcement learning}{The study of learning a distribution over interventions for treatment planning}
\entry{Rewards}{A measurement of value of decisions
taken (e.g. patient survival)}
\entry{Actions}{The set of interventions}
\entry{State}{The collection of patient traits that can inform the next action}
\entry{Policy}{The distribution over interventions given the patient state}
\end{marginnote}

\begin{figure}[t]
    \centering
    \includegraphics[scale=0.35]{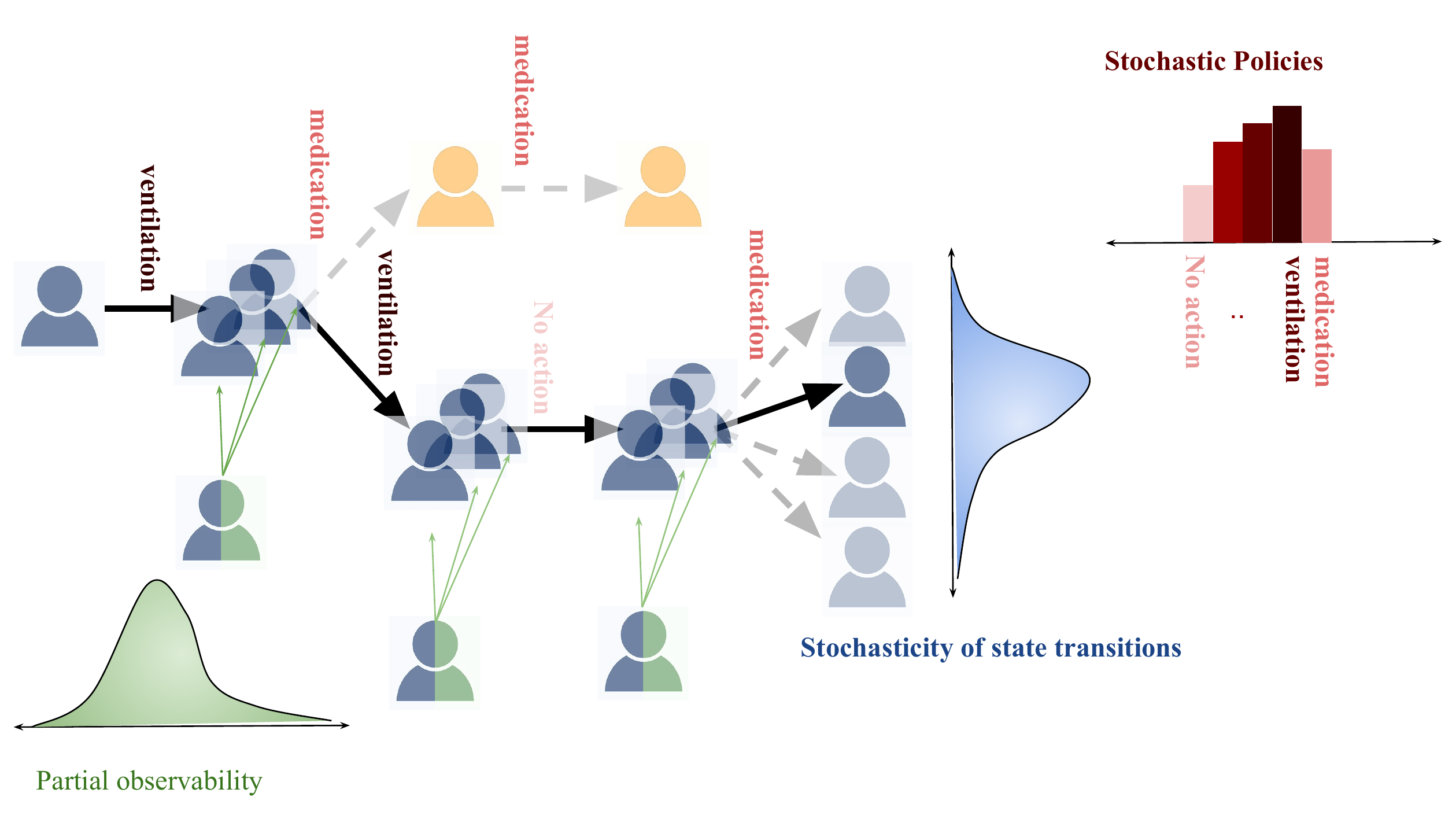}
    \caption{Components of an RL based treatment planning model that should be modeled probabilistically. Patient progression through a treatment plan is stochastic (blue distribution over state transitions) with shades of blue representing possible states. A treatment plan or a policy itself, learned from clinical data and stochastic clinical practice can lead to randomness in interventions creating a distribution over potential interventions (e.g. a patient can be ventilated or given medication or simply be monitored for longer--no-action). This resulting distribution is denoted as a discrete distribution over interventions on the top-right where the shades of red denote the type of intervention. Finally, many aspects of a patient may be unobserved (green+blue shaded patients represent unobserved patient states in green) but critical to treatment planning. Modeling this partial observability (the green distribution denotes this randomness) is beneficial for policy learning.}
    \label{fig:rl_prob}
\end{figure}

\subsection{Model-based RL}
Canonical RL assumes the ability to conduct experiments in the target environment, which allows the learning algorithm to collect samples to learn a reasonable policy, called ``model-free" learning, explored for diabetes management~\cite{javad2019reinforcement} and sepsis management~\cite{raghu2017continuous,komorowski2016markov}. 
In model-free learning, an RL agent does not model the underlying transition dynamics of the data,
which are how a patient state changes based on their current state and the current intervention performed by the clinician. In healthcare, it is not feasible to conduct online experiments, due to ethical concerns around patient safety, neither is it easy to collect vast amounts of data. 
Embedding prior knowledge about transition dynamics is therefore key in healthcare and can reduce the need to collect many samples. This is usually done via ``model-based" RL, which involves explicitly modeling the transition dynamics to learn a policy. 
For example, consider the problem of managing type-1 diabetes, where the goal would be to determine amount of insulin to administer based on continuous glucose monitoring~\cite{foxdeep}. While the underlying physiological glucagon kinetics and secretion are well characterized~\cite{man2014uva}, carbohydrate intake of a patient is stochastic~\cite{foxdeep}. This makes the transition dynamics of the patient's glucose model partially stochastic~\cite{harris1919biometric} and should be modeled accordingly for healthcare RL tasks. 

\subsection{Stochastic Policies}
Clinical data is rife with different forms of uncertainty. The nature of uncertainty can be due to noisy or even incomplete observations of patient state. In Section~\ref{sec:uncertainty}, we focused on capturing uncertainty for predictive and diagnostics tasks. Here, we focus on how uncertainty should be accounted for treatment planning in sequential decision making.
First, there is inherent randomness in how patients are treated as well as in healthcare data-collection. For example, dosages can differ across hospital practices; so can reporting standards and practices~\cite{gottesman2019guidelines,van2016variation}. There may also be some uncertainty between when a drug was administered versus when the intervention was recorded.
Modeling policies deterministically can then lead to misspecification in RL problem formulation. In the extreme case where data-collection and recording artefacts are modeled without any wiggle room to model the randomness, one runs the risk of blind imitation of such artefacts rather than learning clinically meaningful treatment policies. 


Most RL methods (policy learning or evaluation) in healthcare fall in the framework of off-policy RL. Off-policy learning refers to policy learning when observational data on patient progression and interventions are available from a clinician's treatment path, called the \emph{behavior} policy. 
The goal is to either learn a better policy or estimate the reward of a different policy on these patients. Off-policy means the algorithm is unable to interact with the environment (patients) to collect new samples.
Here again one of the main statistical challenge is sample efficiency to learn a better policy. 

Along with modeling the transition dynamics, the inability to collect additional data (by actively intervening on patients) can be mitigated by framing off-policy learning as a causal (and therefore a probabilistic) estimation problem: “would the patient have survived had they not been treated?”
This framing allows the RL algorithm to explore how outcomes could have changed under a slightly different set of interventions without actual experimentation.
Another way to address this is if a similar patient happened to be in the dataset who was not treated and one could similarly estimate a good treatment policy based on the estimated efficacy of the treatment. 
Causal modeling offers an in-between solution, where by making certain assumptions on the underlying probabilistic data-generating mechanisms, one could sample such \emph{counterfactual} trajectories from the original observed data itself~\cite{oberst2019counterfactual}. Thus, with fewer effective number of samples, reliable stochastic policies can be learned with a probabilistic approach to RL. 
In healthcare, the benefits of causal probabilistic modeling have become a complementary toolbox that can be leveraged for training reliable policies.

\subsection{Partial Observability}
In many cases, clinicians' interventions are done with more implicit information than available to an RL algorithm. This problem is known as partial observability and is yet another source of uncertainty in reliable policy learning. The conventional MDP framework can be augmented to handle partial observability, known as Partially Observable MDPs. The added modeling complexity now involves learning an observation function $p(o| s)$, which characterizes likely observations $o$ an RL algorithm perceives based on the potential states $s$ of the patient. 
Confounding in observational health data is one such example source of partial observability~\cite{gottesman2019guidelines}. 
Consider a case where socio-economic status is unavailable to an RL agent to learn from but was used by clinicians, who may have used costlier treatments for wealthy patients. The goal is then to learn a policy when a behavior policy operated on more (and complete \emph{state} information) compared to what is available to the RL algorithm. In this case, learning with partial observations involves estimating the posterior over unobserved states $p(s | \cdot)$~\cite{sallans2000learning,tennenholtz2019off}. The benefits of such probabilistic inference have been demonstrated for off-policy evaluation in the presence of such confounding for treatment planning~\cite{tennenholtz2019off}.

\section{DISCUSSION}
Machine learning for healthcare holds promise to reshape healthcare. In this review, we give a
broad overview of the role that probabilistic modeling plays in medicine. 
This review does not touch on all of the areas of healthcare that benefit from 
probabilistic modeling. For example, causal inference \cite{hernan2020causal} is a central question in medicine. Probabilistic methods 
have been used to estimate causal effects in HIV treatment~\cite{cain2010start}, and in general, probabilistic techniques have been shown to give some of
the most accurate inferences at a well-known causal inference challenge \cite{dorie2019automated}. As another example, healthcare often involves
time-series data, which benefits from a probabilistic
approach for providing uncertainty estimates over forecasts~\citep{krishnan2017structured}.

We believe that a probabilistic perspective can yield significant benefits when building machine learning models for healthcare;
this review covers specific examples in building predictive models, phenotyping, sample generation, and learning policies. 
We encourage further research into both methodologies and applications with probabilistic machine learning models in healthcare.

\section*{DISCLOSURE STATEMENT}
The authors are not aware of any affiliations, memberships, funding, or financial holdings that
might be perceived as affecting the objectivity of this review. 

\section*{ACKNOWLEDGMENTS}
The authors thank No\'{e}mie Elhadad, Rahul G. Krishnan, Peter Schulam, and Pete Szolovits for helpful and useful feedback. This work was supported in part by a CIFAR AI Chair at the Vector Institute (MG) and Microsoft Research (MG).

\bibliographystyle{ieeetr}
\bibliography{references}

\end{document}